\documentclass[conference, final]{IEEEtran}

\usepackage{amsmath,amssymb,amsfonts}
\usepackage{algorithmic}
\usepackage{graphicx}      
\usepackage[inkscapelatex=false]{svg}
\usepackage{textcomp}
\usepackage{xcolor}
\usepackage{hyperref}
\usepackage[noadjust]{cite}

\usepackage{scalerel}
\usepackage{tikz}
\usetikzlibrary{svg.path}
\definecolor{orcidlogocol}{HTML}{A6CE39}
\tikzset{
  orcidlogo/.pic={
    \fill[orcidlogocol] svg{M256,128c0,70.7-57.3,128-128,128C57.3,256,0,198.7,0,128C0,57.3,57.3,0,128,0C198.7,0,256,57.3,256,128z};
    \fill[white] svg{M86.3,186.2H70.9V79.1h15.4v48.4V186.2z}
                 svg{M108.9,79.1h41.6c39.6,0,57,28.3,57,53.6c0,27.5-21.5,53.6-56.8,53.6h-41.8V79.1z M124.3,172.4h24.5c34.9,0,42.9-26.5,42.9-39.7c0-21.5-13.7-39.7-43.7-39.7h-23.7V172.4z}
                 svg{M88.7,56.8c0,5.5-4.5,10.1-10.1,10.1c-5.6,0-10.1-4.6-10.1-10.1c0-5.6,4.5-10.1,10.1-10.1C84.2,46.7,88.7,51.3,88.7,56.8z};
  }
}

\newcommand\orcidicon[1]{\href{https://orcid.org/#1}{\mbox{\scalerel*{
\begin{tikzpicture}[yscale=-1,transform shape]
\pic{orcidlogo};
\end{tikzpicture}
}{|}}}}

\hypersetup{
    colorlinks=true,
    linkcolor=blue,
    citecolor=blue,
    filecolor=blue,      
    urlcolor=cyan,
    pdftitle={},
}

\begin{document}

\title{A Grid Cell-Inspired Structured Vector Algebra for Cognitive Maps}

\author{
\IEEEauthorblockN{Sven Krausse \orcidicon{0009-0001-9997-9022}}
\IEEEauthorblockA{\textit{Forschungszentrum Jülich} \\
\textit{RWTH Aachen}\\
Aachen, Germany \\
s.krausse@fz-juelich.de}
\and
\IEEEauthorblockN{Emre Neftci \orcidicon{0000-0002-0332-3273}}
\IEEEauthorblockA{\textit{Forschungszentrum Jülich} \\
\textit{RWTH Aachen}\\
Aachen, Germany \\
e.neftci@fz-juelich.de}
\and
\IEEEauthorblockN{Friedrich T. Sommer \orcidicon{0000-0002-6738-9263}}
\IEEEauthorblockA{\textit{UC Berkeley} \\
\textit{Redwood Center for} \\
\textit{Theoretical Neuroscience}\\
Berkeley, USA \\
fsommer@berkeley.edu}
\and
\IEEEauthorblockN{Alpha Renner \orcidicon{0000-0002-0724-4169}}
\IEEEauthorblockA{\textit{Forschungszentrum Jülich} \\
Aachen, Germany \\
a.renner@fz-juelich.de}
}

\maketitle

\begin{abstract}
The entorhinal-hippocampal formation is the mammalian brain's navigation system, encoding both physical and abstract spaces via grid cells. This system is well-studied in neuroscience, and its efficiency and versatility make it attractive for applications in robotics and machine learning.

While continuous attractor networks (CANs) successfully model entorhinal grid cells for encoding physical space, integrating both continuous spatial and abstract spatial computations into a unified framework remains challenging.
Here, we attempt to bridge this gap by proposing a mechanistic model for versatile information processing in the entorhinal-hippocampal formation inspired by CANs and Vector Symbolic Architectures (VSAs), a neuro-symbolic computing framework. The novel grid-cell VSA (GC-VSA) model employs a spatially structured encoding scheme with 3D neuronal modules mimicking the discrete scales and orientations of grid cell modules, reproducing their characteristic hexagonal receptive fields.

In experiments, the model demonstrates versatility in spatial and abstract tasks: (1) accurate path integration for tracking locations, (2) spatio-temporal representation for querying object locations and temporal relations, and (3) symbolic reasoning using family trees as a structured test case for hierarchical relationships.

\end{abstract}


\begin{figure*}[btp]
\centerline{\includegraphics[width=0.92\linewidth]{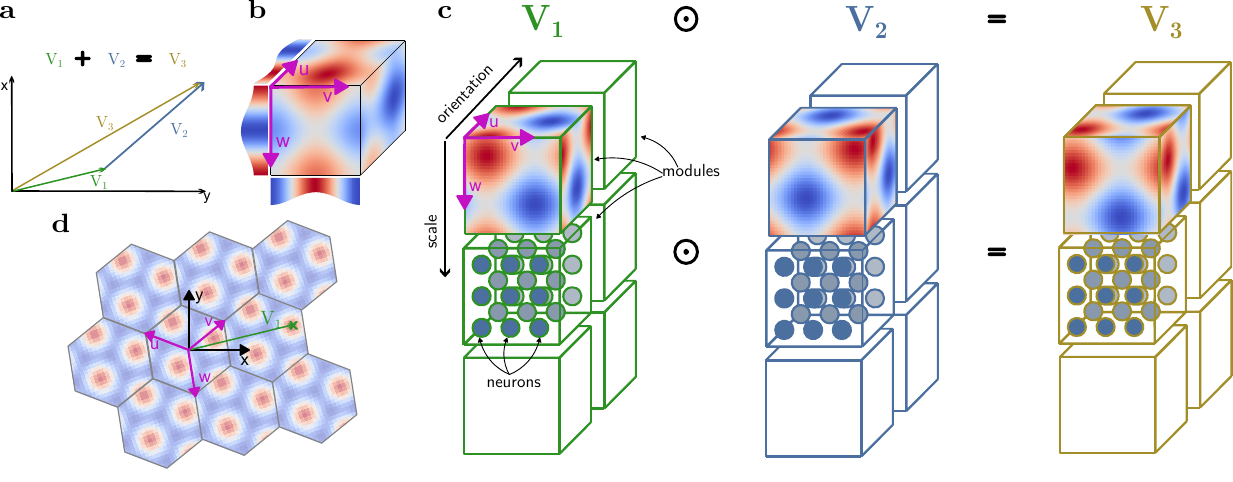}}
\caption{\textbf{Schematic of the proposed GC-VSA.} 
\textbf{a}: Geometric analogy of the binding operation with FPE encoded vectors: Here, binding corresponds to vector addition in 2D Euclidean space.
\textbf{b}: Visualization of how the 3D module activation is constructed from the outer sum of 3 cosines. Depending on the three phases along the three axes $u,v,w$, a cosine grating with the corresponding phase shift is used along the corresponding axis, as indicated by the heatmap at each of the three sides of the cube. The activity at each point in the cube is then computed as the outer sum of these three cosine gratings.
\textbf{c}: Schematic of the encoding scheme and the binding operation. Encoding tensors consist of a set of cubes with varying orientations and scales (i.e., relative angles and vector lengths between the hexagonal (pink) and cartesian (black) coordinate system in panel d). Each tensor consists of $n_\theta \times n_s$ modules ($n_\theta=2, n_s=3$ is visualized here, $n_\theta=23, n_s=5$ in the simulations), and each module consists of $n \times n \times n$ neurons (here $n=3$). The activity pattern of each module is shown in panel b. This activity pattern is discretized by an equidistantly spaced grid of the $n \times n \times n$ neurons, as visualized  in the middle modules for each vector. The binding operation functions as a circular convolution for each module of the same scale and orientation. This can be seen in the activity patterns of the three modules, where $V_3$ is the activity pattern resulting from a 3D circular convolution of the activity pattern of $V_1$ by the activity pattern in $V_2$. 
\textbf{d}: Mapping of 2D space with one periodic 3D module. The image shows the 2D projection of the activity of a single 3D module, as shown in panel b. The three axes of the cube ($u,v,w$, pink) in panel b are projected to 3 hexagonal axes in 2D space that are $120^\circ$ apart. Each hexagon shows an identical repeated activity pattern (as it is derived from the same module). The scale and orientation of the hexagonal coordinates ($u,v,w$) with respect to the cartesian coordinates ($x,y$) depends on the scale and rotation of the 3D module we use to project onto the 2D plane. Intuitively, to encode position $V_1$ (green), the activity pattern is shifted to have a bump at the tip of the vector. When this is done for all modules, they constructively interfere at this location and destructively at other locations (see Fig.~\ref{fig:phase_kernel_explainer}c)~\cite{mcnaughton_2006}.}
\label{fig:encoding}
\end{figure*}

\section{Introduction} 
The entorhinal-hippocampal formation (EHF) is the mammalian brain's navigation system and a crucial hub for memory formation. At its input, grid cells \cite{hafting_2005} in the entorhinal cortex (EC) encode space using a sparse distributed population code organized into modules of different spatial scales and orientations \cite{stensola_2012}. Grid cells exhibit a characteristic periodic hexagonal grid receptive field.

The well-studied entorhinal grid cell system has been modeled using Continuous Attractor Networks (CAN) \cite{mcnaughton_2006, burak_2009, dong_2024} and Oscillatory Interference Models \cite{burgess_2007,hasselmo2007grid}. CANs explain the periodic hexagonal activity pattern by neural activity `bumps' that move equivariantly with the animal on several neural sheets of toroidal topology, called modules. At first glance, a totally different model, the Oscillatory Interference Model, assumes a set of Velocity-Controlled Oscillators (VCO), i.e., oscillators with a frequency dependent on the animal's current velocity. The interference of at least two VCOs tuned to different movement directions gives rise to periodic firing patterns in running animals.  

Curiously, neural responses in the EHF extend beyond physical space. While hippocampal place cells double as `concept cells' \cite{quiroga2009explicit}, grid cells have been shown to encode abstract spaces \cite{constantinescu_2016,whittington_2020,theves2020hippocampus,liang2024distance}, suggesting a general mechanism for organizing spatial information in the brain.
However, a unified computational framework for encoding both physical and abstract spaces remains an open challenge.

In this paper, building on work linking grid cells to Hyperdimensional Computing (HDC) \cite{frady_2018,komer2020thesis,frady_2021,kymn2024binding}, we propose a framework that bridges the gap between physical and abstract spatial encoding strategies. HDC~\cite{plate_1995,kanerva_2009,frady_2018,frady_2021} was designed for symbolic computations in neural systems, enabling the representation of complex data structures like variables, trees, and graphs. Recent studies have extended HDC to continuous spatial variables and manifolds \cite{frady_2021}, making it a promising framework for bridging spatial and symbolic computation. Its robustness, scalability, and interpretability also make it suitable for neuromorphic computing \cite{kleyko_2022,karunaratne2020memory,renner2022sparse,renner_2024}.

HDC is an umbrella term for many different Vector Symbolic Architectures (VSA). A VSA defines an algebraic structure consisting of a set of very high-dimensional (`hyperdimensional') vectors and the two binary commutative and associative operations bundling and binding.

We introduce `grid-cell VSA' (GC-VSA), a variant of Fourier Holographic Reduced Representations (FHRR) \cite{plate_1995}. GC-VSA uses a 3D block-code with additional structure inspired by grid cell organization and defines associated binding and bundling operations. Unlike earlier VSAs  \cite{frady_2018,komer2020thesis,frady_2021,kymn2024binding}, which observe grating patterns in single elements, GC-VSA generates single-neuron responses with hexagonal grid receptive fields, aligning more closely with biological observations and providing an intuitive link between VSAs and neural activity.

To demonstrate the viability of the model, we reproduce the most important proof-of-principle experiments that have been performed in the previous literature  \cite{frady_2018,komer2020thesis,frady_2021}: path integration and spatio-temporal scene/object representation. In addition, the code's structure allows us to introduce a rotation operation. To highlight that the model is capable of abstract spatial reasoning, we demonstrate a simple analogical reasoning task on a tree of family relationships.

Finally, we discuss how the proposed vector algebra relates to previous HDC models and computational neuroscience models. Interestingly, this framework not only unifies aspects of CAN and VCO models but also describes representations and computations in EC that can express spatial and symbolic information processing in a single framework. 

\section{Vector Symbolic Architectures}
\label{sec:methods}
The entorhinal grid cell system is hypothesized to function as a scaffold for encoding and navigating both physical and abstract spaces \cite{mcnaughton_2006,constantinescu_2016,chandra_2023,liang2024distance}. It comprises multiple neuron modules with periodic receptive fields in a hexagonal lattice at different spatial frequencies. Within each module, cells share grid spacing and orientation but differ in spatial phase, ensuring continuous coverage of space \cite{stensola_2012}. In rats, typically 4-5 modules \cite{stensola_2012} operate as CANs \cite{mcnaughton_2006, burak_2009,yoon2013specific, gardner_2022, gu2018map}, with grid spacing increasing in discrete steps along the dorsal-ventral axis of the medial entorhinal cortex (MEC) \cite{stensola_2012}. 
Inspired by this biological architecture, we propose a VSA, which we call grid cell VSA (GC-VSA), which integrates concepts from FHRR \cite{plate_1995} and sparse block codes \cite{laiho_2015}.

Before introducing the details of the proposed model, we provide a quick tutorial on vector symbolic architectures as algebraic structures and how they can be used to represent categorical and spatial data.

\subsection*{Algebraic Structure of VSAs}

Vector Symbolic Architectures (VSAs) are algebraic frameworks that represent information using high-dimensional vectors and define two binary operations: bundling ($\oplus$) and binding ($\odot$). 
The vectors and these operations are implemented differently in different VSAs.
Before detailing their implementation, we outline their algebraic structure.

A fundamental principle of VSAs is that the entire state of a system is represented in a distributed fashion in a single hyperdimensional vector of fixed size. This vector is typically constructed by randomly sampling vectors assigned to different symbols in a codebook matrix, ensuring that they are nearly orthogonal and thus distinct in cosine similarity. For example, vectors can represent categorical information like $banana$ and $apple$.

The binding operation ($\odot$) is classically used to combine features. For instance, we could represent a green apple in the neural state by binding the two vectors labeled $green$ and $apple$: $green \odot apple$. Notably, the binding operation is designed such that it is invertable, i.e. there exists an inverse operation, called unbinding $(\odot^{-1})$. In the example above, unbinding $green$ from the $green$ $apple$ vector returns the vector apple: $(green \odot apple) \odot^{-1} green = apple$.

The bundling operation ($\oplus$) is classically used to represent a set of objects in such a vector. For instance, a scene containing a green apple and a yellow banana is encoded as $S = (green \odot apple) \oplus (yellow \odot banana)$. Crucially, the binding and unbinding operations are associative over the bundling operation. Unbinding $green$ from the scene vector $S$ results in the following vector: 
\begin{equation}
\begin{split}
     &((green \odot apple) \oplus (yellow \odot banana)) \odot^{-1} green \\
     &=apple \oplus (yellow \odot banana \odot^{-1} green).
\end{split}
\end{equation}
The resulting bundle is most similar to $apple$ because the second term merely adds noise as it is not similar to any of the vectors in the codebook with which the readout is performed.
In this example, unbinding of $green$ corresponds to asking the question 'Which of the fruits in the scene is green?'.

Note that binding and bundling preserve the vector size, maintaining constant memory usage. However, the representation quality degrades with excessive bundling.

Beyond categorical representation, as presented in this section, VSAs can encode continuous spatial variables, supporting geometric operations.

\subsection*{Spatial Representation with VSA}

To extend these operations to spatial information, the binding operation can be applied repeatedly to represent integer and real numbers~\cite{plate_1995}. We write $x$ repetitions of self-binding of the generator vector $V_X$ as eponentiation: $V_X^{\odot x}$~\cite{renner_2024}. This allows the encoding of integer numbers ($\mathbb{Z}$) by labeling $V_X$ as 1, $V_X \odot V_X$ as 2, $V_X^{\odot 3}$ as 3, and so forth. 
By binding with another random generator vector $V_Y$, we can span a 2D space, allowing the representation of coordinates in $\mathbb{Z}^2$. For example, the coordinate (3,7) is represented by $V_X^{\odot 3} \odot V_Y^{\odot 7}$. This binding operation gives us the means to navigate in this space, enabling movements such as shifting 2 units in the $x$-direction and 1 unit in $y$, computed as follows: $V_X^{\odot 3} \odot V_Y^{\odot 7} \odot V_X^{\odot 2} \odot V_Y^{\odot 1} = V_X^{\odot 5} \odot V_Y^{\odot 8}$. 

I.e., the binding operation in the hyperdimensional space represents vector addition in the low-dimensional, encoded space:
\begin{equation}
    V_X^{\odot x_1} \odot V_X^{\odot x_2} = V_X^{\odot (x_1 + x_2)}
\end{equation}

This feature is exactly what we will need for path integration in the next section, but so far, we are limited to integer steps on a lattice. To represent real vectors (in $\mathbb{R}^2$) we use fractional power encoding (FPE) \cite{plate_1995,frady_2019,komer_2019}, where the binding operation can be applied fractionally, allowing the exponents $x$ and $y$ to take real values:
\begin{equation}\label{eq:pos_encoding}
  V_{x,y} = V_X^{\odot x} \odot V_Y^{\odot y}.
\end{equation}
Intuitively, fractional binding corresponds to partial (non-integer) shifts enabling smooth interpolation between discrete encoded positions.
Importantly, FPE preserves locality, i.e. vectors close in physical space (with exponents that differ by less than 1) are similar in hyperdimensional space as measured by cosine similarity. The similarity kernel in 2D is depicted in Fig.~\ref{fig:phase_kernel_explainer}c. 

An ideal VSA for FPE is the Fourier Holographic Reduced Representation (FHRR) \cite{plate_1995}. In FHRR, basic hyperdimensional vectors are encoded as complex numbers of amplitude 1, so-called phasors. The bundling operation is implemented as an element-wise addition, and the binding operation is an element-wise multiplication. Thus, repeated self-binding is implemented as elementwise exponentiation, which can easily be generalized to non-integer exponents.
For phasors, complex multiplication is phase addition, so continuous movement through the represented space leads to rotations of the complex phases (i.e. oscillations, providing a link to VCOs).

\subsection*{Block-Coded VSAs}
As a complex number describes a cosine of a certain phase and magnitude, an alternative implementation of FHRR is possible:
Here, we describe a VSA of high-dimensional vectors that are structured in multiple blocks of real values: $V=[v_1, ..., v_m]$, where each block \(v_i \in \mathbb{R}^{n}\) encodes a discretized cosine function with a specific phase and magnitude corresponding to a single element in FHRR.

Binding in this encoding is performed via block-wise circular convolution. Since circular convolution is equivalent to elementwise multiplication in the Fourier domain, the connection to FHRR’s complex representation can be made explicit using the Fourier transform $\mathbf{F}(\cdot)$:
\begin{equation}
    \label{eq:Fourier_theorem_cc}
    \begin{split}
        W &= U \odot V \\
        \Rightarrow w_i &=\mathbf{F}^{-1}[\mathbf{F}(u_i) \cdot \mathbf{F}(v_i)].
    \end{split}
\end{equation}
The inverse operation, unbinding ($\odot^{-1}$), is performed via a block-wise circular convolution in the inverse direction, i.e., multiplication of the complex conjugate in Fourier space:
\begin{equation}
    \label{eq:unbinding}
    \begin{split}
        W &= U \odot^{-1} V \\
        \Rightarrow w_i &=\mathbf{F}^{-1}[\mathbf{F}(u_i) \cdot \overline{\mathbf{F}(v_i)}].
    \end{split}
\end{equation}

Eqs. \ref{eq:Fourier_theorem_cc} and \ref{eq:unbinding} highlight that both complex vectors and block-wise encoding use the same principle and implement the same VSA (FHRR): In the complex setting, elements are in the Fourier domain while this block code represents each element in the space domain. 
Applying this property enables straightforward implementation of fractional power encoding (FPE), where repeated binding corresponds to exponentiation in the Fourier domain. 

This block-wise vector encoding is closely related to sparse block codes (SBC) \cite{laiho_2015}. Sparse block codes also partition vectors into blocks but employ one-hot encoding rather than cosine functions. 

Sparse block codes, however, behave differently from FHRR as one-hot blocks only represent discrete phases without magnitudes. Consequently, bundling leads to multiple activations per block, affecting sparsity. Additionally, using FPE via circular convolution on SBCs disrupts their binary nature. The discretized cosine encoding can therefore be viewed as a continuous extension of SBC. An alternative continuous extension was recently proposed \cite{hersche2024factorizers}.

In the following, we introduce another novel variant of the block-FHRR where each element/block has multiple phases and one magnitude, instead of a single phase and a magnitude.

\section{The Grid-Cell-VSA Model}
FHRR represents symbols as high-dimensional complex-valued vectors. We extend this by replacing complex vector elements by 2D or 3D modules. Here, we only focus on the 3D case where each module is an $n\times n\times n$ tensor ($n=3$ in our experiments).

A grid-cell VSA (GC-VSA) `vector' consists of multiple modules organized by scale and orientation (Fig.~\ref{fig:encoding}c). Accordingly, full GC-VSA `vectors' are 5D tensors with three spatial dimensions for each module, the fourth for orientation of the modules, and the fifth for scale of the modules. Despite this structure, we refer to the entire 5D tensor as a `VSA vector' since it can be flattened into a conventional hyper-dimensional vector. 

\subsection*{Space Encoding with GC-VSA}
Following the logic of FPE, each module encodes a given point in 2D-Cartesian space $(x,y)$ by first applying a rotation ($R$) to the module-specific orientation angle $\theta$, then projecting it ($T$) into a hexagonal (3D) coordinate system, and scaling it to module scale $s$ (the orientations and scales are visualized in Fig.~\ref{fig:phase_kernel_explainer}a and b): 
\begin{equation}
\begin{split}
    &\begin{pmatrix} 
        u\\ v \\ w
    \end{pmatrix} = \frac{1}{s} \cdot T \cdot R 
    \begin{pmatrix} 
        x\\ y 
    \end{pmatrix} \\
    &= \frac{1}{s} \cdot 
    \begin{pmatrix}  
        \frac{\sqrt{3}}{2}   & -\frac{1}{2} \\
        -\frac{\sqrt{3}}{2}  & -\frac{1}{2} \\
        0                    & 1
    \end{pmatrix} \cdot 
    \begin{pmatrix}  
        \cos(\theta) & -\sin(\theta) \\
        \sin(\theta) & \cos(\theta) 
    \end{pmatrix} \cdot 
    \begin{pmatrix} 
        x\\ y 
    \end{pmatrix},    
\end{split}
\label{eq:encoding}
\end{equation}
Fig.~\ref{fig:encoding}d shows the two coordinate systems of this transformation: the 2D cartesian coordinate system (black) and the 3D hexagonal coordinate system (pink). 
This projection corresponds to previous hyperdimensional encoding approaches \cite{komer2020thesis,kymn_2023}.
Each module's activation is generated as the outer sum of phase-shifted harmonic cosine functions based on the transformed point $(u,v, w)$, as visualized in Fig.~\ref{fig:encoding}b:
\begin{equation}
\begin{split}
    n_{ijk}=\sigma(&\cos(\frac{2\pi}{n}(i-u)) + \cos(\frac{2\pi}{n}(j-v))\\&+\cos(\frac{2\pi}{n}(k-w))),
\end{split}
\label{outer_sum}
\end{equation}
where $\sigma$ normalizes the module's activation to ensure an amplitude of 1 in the 3D discrete Fourier transform and n is the number of neurons per dimension of the cube module.

Consequently, unlike regular FHRR, where each element encodes a single phase and a magnitude, each GC-VSA module contains a 3D phase (the phases of the cosines) and magnitude. As in FHRR, the magnitude is 1 for basic vectors but changes upon bundling.

\subsection*{Choice of parameters for the GC-VSA encoding}
For the presented experiments, we choose the number of neurons per module dimension $n=3$, as this is the minimal number of sampled points that allows the shift using circular convolution. The $n_{\theta}$ different orientations $\theta$ are distributed uniformly in the range between $0$ and $2\pi$ with a random offset per scale to improve coverage of orientation-scale pairs ($n_{\theta}=23$ in presented experiments). For the different scales, we take inspiration from the neuroscience literature. The different scale modules in entorhinal cortex are known to increase exponentially with a constant factor of 1.42 between them \cite{stensola_2012}, starting from a minimal scale $s_{min}$:
\begin{equation}
    s_i = s_{min} \cdot 1.42^i,
\end{equation}
with $i\in[0,1,...,n_s-1]$ and $n_s$ the number of scales. We use $n_s=5$ scales (following Stensola et al. \cite{stensola_2012}) and set $s_{min}=4$ $pixels$.

\subsection*{Bundling and Binding Operations in GC-VSA}
In GC-VSA, bundling ($\oplus$) is defined as the element-wise addition of neural activations. Binding ($\odot$) is a module-wise circular convolution of the two vectors, shifting one spatial phase by the other and corresponding to multiplication in Fourier space. This generalizes the block-code, described earlier, to 3D (see Eq.~\ref{eq:Fourier_theorem_cc} and ~\ref{eq:unbinding}). Fig.~\ref{fig:encoding}b illustrates how binding vectors $U$ and $V$ results in shifted spatial phases within the 3D modules. Analogous to fractional binding \cite{plate_1995,frady_2018,komer_2019} in FHRR, GC-VSA allows encoding of real-valued positions via fractional circular convolution.

\begin{figure}[htbp]
\centerline{\includegraphics[width=\linewidth]{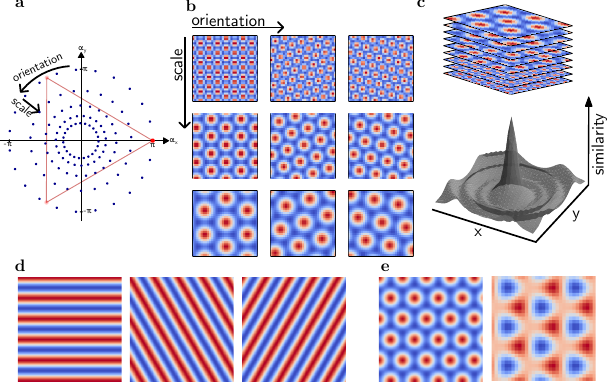}}
\caption{\textbf{The generator phases that span the encoding produce different receptive fields in different VSAs.} 
\textbf{a}: Generator phases determining spatial frequencies across modules (see Fig.1 of Frady et al. \cite{frady_2021}). 
Each point describes one module with a specific scale and orientation. The scale and orientation or spatial frequency of a module is defined by the two spatial phases of the two (or three) generator phases that span the space. In the context of 2D space representation, they can, however, also be regarded as a feature of each module of the structured encoding scheme. 
The three red dots are drawn to remind us that each blue dot actually corresponds to a triplet of generator phases, one for each of the three axes.
\textbf{b}: Example basis functions or receptive fields for different modules. Each basis function shows a hexagonal grid pattern \cite{mcnaughton_2006}. The periodicity and rotation of the pattern are defined by each module's scale and orientation, as shown in \textbf{a}.
\textbf{c}: Intuition for the cosine similarity kernel by stacking and adding the periodic basis functions \cite{mcnaughton_2006}. The constructive interference in the center produces a peak in the form of a 2D sinc function. 
\textbf{d}: Three example gratings that describe receptive fields of FHRR phasors with FPE encoding. Only the addition of 3 complex neurons yields hexagonal receptive fields. 
\textbf{e}: Resulting receptive fields through the addition of three cosine gratings. Depending on the phase relationship between the three encodings, the receptive field of a neuron in the block-code is either of the two shown periodic patterns. Both types of receptive fields are found in each module for different neurons.}
\label{fig:phase_kernel_explainer}
\end{figure}

This relies on generator vectors (e.g. $V_X$ and $V_Y$ in Eq~\ref{eq:pos_encoding}), which are iteratively bound to span the encoding space. The generator phase determines the rate of spatial phase shifts, setting the neuron receptive field frequency (Fig.~\ref{fig:phase_kernel_explainer}a,b). Lower generator phases correspond to slower shifts and, thus, lower spatial frequencies.

When we move in 2D, activation shifts within the 3D modules at three different velocities, arranged $120^\circ$ apart (purple arrows in Fig.~\ref{fig:encoding}d). This triplet of velocities is depicted as a red triangle in Fig.~\ref{fig:phase_kernel_explainer}a. Unlike prior methods, where grating patterns emerge from independently encoded dimensions, GC-VSA inherently encodes these triplets within 3D modules.

To introduce additional structure into the encoding, we can sort and arrange the modules by generator phases, organizing them as a 5D tensor along the orientation and scale dimensions. Orientations are evenly distributed (although this may differ from biological organization of orientations in EC \cite{stensola_2012}), which allows us to introduce rotations, as described in the next section.

\subsection*{Rotation Operation in GC-VSA}

The structured encoding of GC-VSA enables a novel VSA rotation operation ($\otimes$), corresponding to rotation of an image around a central point.
The operation can be understood as a permutation of the module activity along the rotation axis, i.e. along the rings that share the same scale as visualized in Fig.~\ref{fig:phase_kernel_explainer}b. To get an intuition, think of shifting the activity in the first column of cube modules in Fig.~\ref{fig:encoding}a into the second column of cubes. Beyond discrete permutation, we can also perform a circular convolution on each ring (of all neurons with the same indices, except along the rotation axis) with a vector $V_R$ containing a localized activity bump at the desired angle:
\begin{equation}
  V_R(i,j,k,l,m) = 
  \begin{cases}
    1, & \text{if } l=1 \\
    0, & \text{otherwise } 
  \end{cases}.
  \label{eq:rotation_base}
 \end{equation}
To rotate any encoded 2D position $V = V_X^{\odot x} \odot V_Y^{\odot y}$ by an angle $\alpha$ we apply FPE along the rotation axis:
\begin{equation}
    V_{rot} = V \otimes V_R^{\otimes \alpha \cdot \frac{n_{\theta}}{2\pi}}, \label{eq:rotation}
\end{equation}
where $\frac{n_{\theta}}{2\pi}$ ensures the convolution is normalized by the number of rotation modules $n_{\theta}$. Note, however, that rotation is limited to vectors representing 2D locations and requires a sufficiently dense  generator phase sampling on each ring (in our rotation experiments, we choose $n_{\theta}=23$). Consequently, rotation is limited to locations close to the rotation center, a common limitation of rotations in VSA space that also appears when mapping between vectors in Cartesian and log-polar space \cite{renner_2024}.

\begin{figure}[htbp]
\centerline{\includegraphics[width=\linewidth]{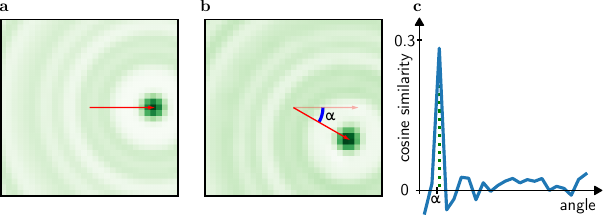}}
\caption{\textbf{Rotation making use of the structure of the GC-VSA encoding.} \\
\textbf{a}: A single 2D position is encoded into a hyperdimensional vector using FPE. The green color shows the similarity of the encoded vector with the codebook vector at each location, and the red vector shows the encoded point. 
\textbf{b}: Using the rotation operation described in Eq.~\ref{eq:rotation}, the vector of panel \textbf{a} is rotated by angle $\alpha$. The true rotated vector is shown in red, and the green color again shows the similarity of the VSA vector to the location codebook. 
\textbf{c}: Rotation decoding: by circular convolution along the rotation axis of the rotated vector with the flipped (analogous to the complex conjugate) version of the other vector, one can read out the angle between the vectors. Here, the rotation vector $\alpha$ is approximately recovered. The green dashed line shows the readout angle.}
\label{fig:rotation}
\end{figure}

Fig.~\ref{fig:rotation} demonstrates encoding a vector using FPE and rotating it by an angle $\alpha$. The rotation operation can also be inverted, and we can decode the angle between two vectors or even bundles of vectors. This is achieved by introducing a rotational unbinding operation ($\otimes^{-1}$) defined analogous to Eq.~\ref{eq:unbinding}. However, here, the Fourier transform is not computed blockwise, but on the ring along each rotation axis. Flattening the resulting tensor onto the rotation axis results in a readout of the angle between the two vectors as shown in Fig.~\ref{fig:rotation}c.

\subsection*{Decoding Vectors using Codebooks}
To decode hyperdimensional vectors, we define a codebook $C$ of possible vectors. For categorical encodings, the codebook consists of the set of all possible categorical encodings. For continuous values, we use the base vector of the continuous variable and encode a range of values using FPE at some specified stepsize (see above). The readout is computed by measuring the cosine similarity between the target vector $V$ and all vectors in the codebook $C_i$:
\begin{equation}
    similarity(C_i, V) = \frac{C_i \cdot V}{\left\lVert C_i\right\rVert \left\lVert V\right\rVert}
\end{equation}
We then finally obtain the decoded value by obtaining the key of the codebook vector with the maximal similarity to the target vector V.

\subsection*{Software}
The simulations were implemented in Python 3.12 using PyTorch (2.4.0), NumPy (1.26.4), and Matplotlib (3.8.4). Grammar and style of the manuscript were revised using Grammarly and OpenAI ChatGPT (4o).

\section{Results}

This section presents experiments demonstrating the GC-VSA's versatility in spatial and abstract tasks. These include (1) path integration, (2) spatio-temporal object representation, and (3) symbolic computation on trees of family relations, bridging HDC’s symbolic capabilities with the spatial encoding of grid cells.

\subsection*{Path Integration}

Path integration (PI) is a critical function of EC~\cite{mcnaughton_2006,burak_2009}, enabling the tracking of an animal’s location by integrating velocity inputs primarily from the vestibular system (inertial measurement). Here, we demonstrate PI using the GC-VSA with FPE (see Eq.~\ref{eq:pos_encoding}).

\begin{figure*}[ht]
  \centerline{\includegraphics[width=.92\linewidth]{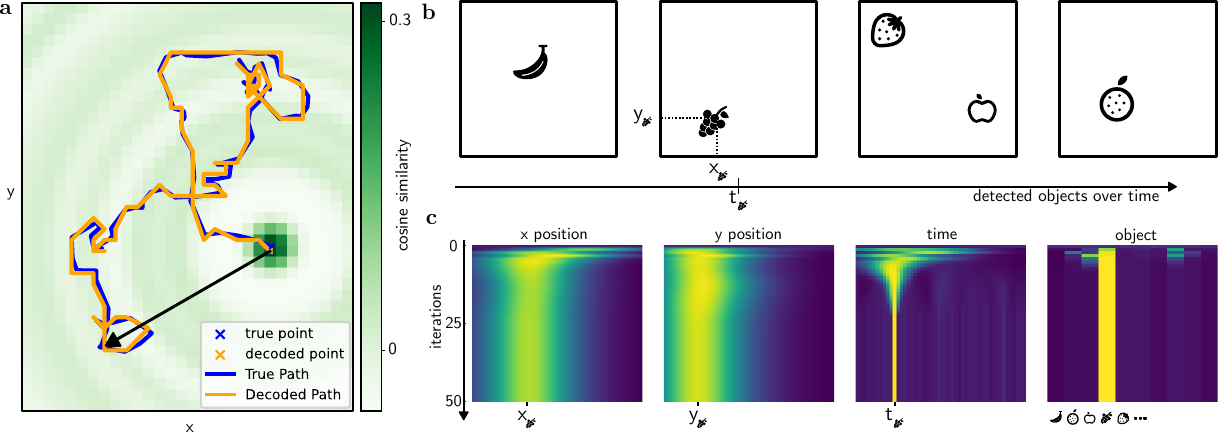}}
  \caption{\textbf{Path integration and Scene representation.} \textbf{a}: 
  Trajectory decoding for path integration. The blue line represents the true path, the orange line shows the decoded trajectory. The green colormap indicates the cosine similarity of the encoded position vector with the codebook. Each vector state encodes the allocentric position, so we can easily get the vector back to the origin (or any other position) at any point in time (here shown as a black arrow).
  \textbf{b}: Visualization of a spatio-temporal scene containing five objects at four points in time. The scene is encoded into a single vector, each object encoded with spatial, temporal, and identity features.
  \textbf{c}: Decoding with the resonator network. Iterative decoding of object position, time and identity from the bundled scene vector. The color describes the similarity between the estimated vector and the codebook vectors over multiple iterations (vertical axis). Yellow color means high similarity and blue low similarity.}
\label{fig:path_integration_and_object_loc}
\end{figure*}

We encode a 2D starting position ($x$,$y$) as a hyperdimensional vector state $S_{x,y}(0)$ using Eq.~\ref{eq:pos_encoding} and then generate a random trajectory (blue in Fig.~\ref{fig:path_integration_and_object_loc}a) by a random series of velocity vectors with autocorrelation. Path integration accumulates velocity updates via repeated binding~\cite{renner_2024a,orchard2024efficient}, where $V_{X,Y}(t)$ encodes displacement at each timestep, according to Eq.~\ref{eq:pos_encoding}:
\begin{equation}
    S_{x,y}(t) = S_{x,y}(t-1) \odot V_{X,Y}(t)
    \label{eq:PI_update}
\end{equation}
The resulting trajectory is decoded by computing the cosine similarity between the VSA vector $S_{x,y}(t)$ with the codebook $C(x,y) = V_X^{\odot x} \odot V_Y^{\odot y}$ for all combinations of $x$ and $y$ at each point in time. 
Peaks in the resulting 2D similarity map (green, Fig.~\ref{fig:path_integration_and_object_loc}a) correspond to the estimated coordinates $(\hat{x}, \hat{y})$.
The reconstructed trajectory (orange, Fig.~\ref{fig:path_integration_and_object_loc}a) closely follows the true path with a mean squared error of 0.17 pixels over 100 timesteps (noise-free condition).

\subsection*{Spatio-Temporal Scene Representation}

Grid cells are not limited to representing 2D spatial locations~\cite{klukas_2020}. We demonstrate the GC-VSA's ability to encode and query a spatio-temporal scene with symbolic components, as visualized in Fig.~\ref{fig:path_integration_and_object_loc}b.

The experiment encodes a scene containing five objects, each with a 2D spatial location $(x, y)$ and a temporal coordinate $t$ as a scaffold, bound to an object identity (e.g., $apple$). The spatial and temporal features are encoded using FPE, while identity is represented by randomly generated vectors (random phase offsets per module). Features of the same object are bound together, and the objects are bundled to form the scene vector:

\begin{equation}
\begin{split}
        map =&       (apple  \odot V_X^{\odot x_a} \odot V_Y^{\odot y_a} \odot V_t^{\odot t_a}) \\
             &\oplus (banana \odot V_X^{\odot x_b} \odot V_Y^{\odot y_b} \odot V_t^{\odot t_b}) \\
             &\oplus (cherry \odot V_X^{\odot x_c} \odot V_Y^{\odot y_c} \odot V_t^{\odot t_c}) \\
\end{split}
\end{equation}
This vector serves as a working memory for scene representation or maps. While GC-VSA supports such dynamic encoding in neural activations, larger scenes require synaptic memory for scalability, such as auto-associative memory.

The encoded scene can be queried to retrieve object identities, locations, or times by unbinding known features. For example, unbinding $apple$ from the scene vector isolates the position and time components for the apple, along with noise from other bundled elements. If three out of four features are known (e.g., $(x, y, t)$), the unknown feature (e.g., identity) is decoded via cosine similarity with a codebook. The cells in the readout of this decoding could be interpreted as place or concept cells, because they are sensitive to a specific location or object identity.

However, when fewer features are known, decoding is challenging due to the large combinatorial space that arises when bound features have to be read out. For instance, there are $D_x \times D_y \times D_t$ (with $D_i$ the number of vectors in the codebook of $i$) possible locations in space-time which can no longer be efficiently covered by pre-computed codebook vectors. 

To tackle this challenge, we employ a resonator network \cite{frady_2020, kent2020resonator, renner_2024} for iterative factorization of bound vectors. The resonator was constructed as in previous work~\cite{renner_2024} but with GC-VSA instead of FHRR. The resonator network iteratively refines initial estimates by repeatedly amplifying consistent features and suppressing inconsistent ones, effectively 'resonating' towards stable solutions. As shown in Fig.~\ref{fig:path_integration_and_object_loc}c, the resonator network decodes object features in under 25 iterations, recovering both spatial and temporal components of an object (e.g., grape) from the bundled scene. Multiple objects could be decoded sequentially by `explaining away' each element \cite{frady_2020,kent2020resonator,renner_2024}, or via multi-headed resonators \cite{renner_2024}.
This experiment illustrates GC-VSA's capacity to integrate spatial and symbolic information in a unified framework.

\subsection*{Spatial-Symbolic Computation - Analogical Reasoning on Trees}
\label{subsec:familytrees}
While grid cells primarily encode physical spaces, evidence suggests they also represent abstract spaces \cite{constantinescu_2016,whittington_2020,liang2024distance}. HDC \cite{plate_1995,kanerva_2009,frady_2018,frady_2021} was designed for implementing symbolic computations in neural systems, which also includes data structures such as trees, and graphs \cite{kleyko_2022,cotteret2025distributed}. Therefore, here, we explore the use of the GC-VSA for representing trees, such as a tree of family relations, inspired by \cite{whittington_2020}. 

\begin{figure*}[ht]\label{fig:here}
   \centerline{\includegraphics[width=\linewidth]{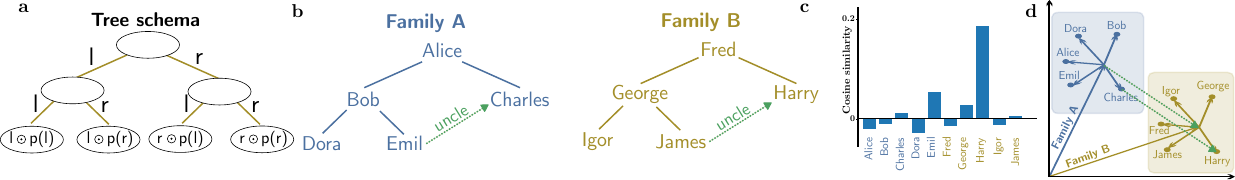}}
   \caption{\textbf{Analogical reasoning on a family tree.}
   \textbf{a}: Encoding scheme of the family tree, using the two base vectors $l$ and $r$ to navigate on a binary tree and using the permutation operation $p(\cdot)$ account for the depth of the tree.
   \textbf{b}: The two encoded trees of family relations. Nodes represent family members, and edges relationships. 
   \textbf{c}: Analogical reasoning: Decoding the analog of 'Charles' from tree $F_A$ in tree $F_B$ using cosine similarity.
   \textbf{d}: Low-dimensional intuition of the spatial computation in the analogy task.
   The green arrow represents the mapping vector $M_{AB}$, applied to identify analogs in the corresponding tree.
   }
 \label{fig:familiytree}
\end{figure*}

We construct a relationship tree where nodes represent family members, and edges encode relationships via binding and permutation operations~\cite{frady_2020}. The tree visualized in Fig.~\ref{fig:familiytree}a is encoded as follows: 

\begin{equation}
\label{eq:tree}
\begin{split}
    F_A = & Alice \oplus (Bob \odot L) \oplus (Charles \odot R) \\
    &\oplus (Dora \odot L \odot (L \otimes V_R)) \\
    &\oplus (Emil \odot L \odot (R \otimes V_R)), \\
\end{split}
\end{equation}
where $L$, $R$, and the names are randomly chosen vectors (i.e., each module has a random offset), $\otimes$ is the permutation operation (here along the rotational axis), and $V_R$ is the base vector for a rotational shift (see Eq. \ref{eq:rotation_base}). A second tree ($F_B$) is constructed analogously with different names. As is typically done when representing binary trees in VSA, the permutation operation is used to counteract the commutativity of the binding operation. The binding operation alone would lead to the same position encoding for 'first left, then right' and 'first right, then left (i.e., $L \odot R = R \odot L$).
Therefore, we need to additionally represent the depth of the tree at which the $L$ or $R$ vector is applied. This is done by repeatedly applying the permutation at each increasing depth of the tree (depth 1 = 0 permutations, depth 2 = 1 permutation, etc.).

The construction now allows us to represent relationships and compute analogies between the two trees.
For instance, Emil could ask James the question, `Who is the Charles in your family?'. The answer to this is computed analogously to the common VSA analogical reasoning tutorial `What is the Dollar of Mexico?' \cite{kanerva_2010}.

We first unbind ($F_A$) from ($F_B$) to derive a mapping vector between the two trees:
\begin{equation}
\begin{split}
    M_{AB} = &F_B \odot^{-1} F_A \\
    = &Fred \odot^{-1} Alice \\
    &\oplus George \odot^{-1} Bob \\
    &\oplus Harry \odot^{-1} Charles \\
    &\oplus Igor \odot^{-1} Dora \\
    &\oplus James \odot^{-1} Emil \\
    &\oplus noise
\end{split}
\end{equation}

The resulting vector $M_{AB}$ corresponds to the green arrow shown in the low-dimensional representation of the task in Fig.~\ref{fig:familiytree}c.
Once computed, $M_{AB}$ captures the structural transformation between the two trees. Applying it to a known node in tree $F_A$ (e.g., $Charles$) retrieves its analog in tree $F_B$ (e.g., $Harry$).
\begin{equation}
Harry \approx Charles \odot M_{AB}
\end{equation}

To read out the result, as shown in Fig.~\ref{fig:familiytree}b, the cosine similarity of the resulting vector with the codebook identifies the most probable match, $Harry$.

\section{Discussion}
This work introduces a vector algebra, GC-VSA, that bridges spatial and symbolic computation by combining hyperdimensional computing (HDC) principles with modular organization observed in entorhinal cortex (EC). 
The framework provides insights into grid cell function in spatial and non-spatial tasks, with possible implications for computational neuroscience and neuromorphic engineering.

\subsection*{Neuromorphic Computing}
HDC has gained traction as an encoding framework in neuromorphic computing due to its interpretability, robustness, and compatibility with parallel, in-memory processing architectures \cite{kleyko_2022,karunaratne2020memory,renner2022sparse,renner_2024}. 

Given GC-VSA’s structured encoding, it is well-suited for neuromorphic implementations. It leverages HDC’s advantages as it is modular, robust, and sparse. The binding and bundling operations are module-local, enabling parallel and scalable implementations. The block-distributed grid cell population code strikes a balance between fully distributed and sparse but inefficient localist space representations \cite{kreiser2018pose,renner2022sparse}. 

Sparse communication via spikes is one of the most crucial features to achieve energy efficiency on neuromorphic hardware. While the state of the neurons in GC-VSA is not sparse (cosine activation pattern), a single spike per module would suffice to roughly readout/communicate and update the state.

Furthermore, the 3D block-code structure enhances sparsity compared to 1D codes: neurons activate only at specific grid points, reducing average firing frequency compared to grating-based receptive fields. This aligns with observations that sparsity constraints yield grid-like receptive fields in learning-based approaches, while grating patterns emerge when the sparsity constraints are removed \cite{gao2021path}.

\subsection*{Relation to Existing VSA Models}
In one of the traditional VSA models, Fourier Holographic Reduced Representation (FHRR) \cite{plate_1995}, symbols and continuous numbers are represented by phasor vectors. This model has been used in earlier models of entorhinal cortex \cite{kymn2024binding}. Our method extends the phasor vector component in FHRR with a single phase by introducing additional phases per element \cite{yeung_2024} and representing these as spatial phases in blocks \cite{laiho_2015}. From a theoretical algebraic perspective, the arrangement of the sine-wave-shaped activity vectors in our model in modules is merely an implementation feature. In contrast, binary sparse block-codes \cite{laiho_2015} have a single binary activation per block, which quantizes the state. In such an encoding, the bundling operation does not preserve sparsity, and fractional binding removes both sparsity and their binary nature. Alternative block-code generalizations have been proposed in the literature \cite{frady_2020a, hersche2024factorizers}.  
In our model, each 3D module can be seen as one component with 3 phases and a magnitude, with the three phases not independent of each other, but at a 120° angle \cite{komer2020thesis,kymn2024binding} which improves the efficiency of the code~\cite{mathis_2015,wei2015principle,komer2020thesis,kymn2024binding}. 
The binding operation, circular convolution, as well as fractional binding shift spatial phases within a module, analogous to complex multiplication in FHRR, which shifts complex phases within an element. 

Previous work has also linked real-valued VSA encodings, the so-called Holographic Reduced Representation (HRR) to grid cell activity in the entorhinal cortex \cite{komer2020thesis}. In this encoding, the binding operation is implemented as a full circular convolution of two vectors. This makes the binding operation expensive. In contrast, our proposed GC-VSA breaks up the binding operation into individual modules, allowing for more efficient computation. Most operations in GC-VSA remain block-local, thereby showing a more bio-plausible mechanism and enabling efficient hardware implementation.

We also propose to impose additional structure to how the vector modules are arranged, which enables rotations to a certain extent. This feature exceeds traditional VSAs and enables spatial computations such as calculating angles between two vectors, for instance, to determine walking direction. The structure along the scale dimension could theoretically be used similarly, for instance, to shift focus from fine to coarse spatial scales.
While previous approaches have introduced the idea of applying different kinds of binding operations to the same VSA vector~\cite{frady2021variable,cotteret2025distributed}, here, we explicitly use it to extend the algebraic properties, i.e. introduce rotations.

\subsection*{Biological Plausibility and Impact to Neuroscience}
The proposed mechanistic model of the EHF bridges
modular assumptions supported by experimental observations in EC with a novel encoding scheme for hyperdimensional computing. 
Previous work has linked VSAs to grid cells  \cite{frady_2018,komer2020thesis,frady_2021,kymn2024binding}. These models typically exhibit hexagonal patterns in the similarity kernel \cite{frady_2021} while the receptive fields of individual neurons are sine-wave gratings. In contrast, the 3D module implementation of the GC-VSA directly exhibits hexagonal fields in individual cells and is more in line with the canonical CAN model as we observe activation bumps that are shifted during path integration instead of shifting of a complex phase.
In the CAN framework \cite{mcnaughton_2006,burak_2009}, each grid cell module is described by a sheet of neurons. The sheet's connectivity/topology is designed to maintain a local activity bump that can be shifted according to head direction and velocity signals. Such bumps of activity moving around with periodic boundary conditions on a (twisted) toroidal topology  \cite{guanella2007model,gardner_2022} give rise to the characteristic hexagonal receptive fields observed in individual neurons.
Here, we demonstrate that the CAN with minor adjustments (specifically adding circular convolution binding instead of just path integration) naturally features the algebraic properties of a VSA. 
Note that, in this work, we focus on the encoding and do not implement a recurrent connectivity. An attractor network could be constructed easily with the Hopfield learning rule, i.e., by forming an auto-associative memory matrix with a module-wise outer product of all locations within a module.

\subsection*{Future Directions}
Our current framework emphasizes working memory without synaptic learning. Future iterations could incorporate machine learning or biologically plausible learning mechanisms, such as hetero-associative connections into the hippocampal CA1 \cite{whittington_2020,chandra_2023,kymn2024binding}. Reinforcement learning and self-supervised paradigms could enable agents to learn spatial and abstract maps dynamically, rather than relying on pre-wired synaptic structure.

Interestingly, the FPE is similar to rotary positional encoding (RoPE) used in transformer architectures~\cite{su2024roformer}, as both leverage complex-valued rotations to encode positions. Future work could explore whether GC-VSA’s structured encoding could enhance transformer-based spatial representations. 

\section{Acknowledgements}
This research is funded by VolkswagenStiftung [CLAM 9C854].
A.R. thanks Alice Collins for valuable insights on circular convolution. 
The authors thank Paxon Frady and Johannes Leugering for inspiring discussions and advice.

\clearpage
\newpage


\end{document}